\documentclass{article} 
\usepackage[usenames,dvipsnames,svgnames,table]{xcolor}
\usepackage{collas2023_conference,times}
\usepackage{easyReview}

\usepackage{caption}
\usepackage{subcaption}
\usepackage{graphicx}
\usepackage{ulem}
\usepackage[misc]{ifsym}

\usepackage{todonotes}
\setuptodonotes{size=tiny, backgroundcolor=orange!20}

\renewcommand{\todo}[1]{}

\usepackage{amsmath,amsfonts,bm,dsfont}

\def\Figref#1{Figure~\ref{#1}}

\def\eqref#1{equation~\ref{#1}}

\def\ineqref#1{Inequality~\ref{#1}}

\def\defref#1{Definition~\ref{#1}}

\def\conref#1{Conjecture~\ref{#1}}

\def\1{\bm{1}}

\DeclareMathAlphabet{\mathsfit}{\encodingdefault}{\sfdefault}{m}{sl}
\SetMathAlphabet{\mathsfit}{bold}{\encodingdefault}{\sfdefault}{bx}{n}

\providecommand{\mc}[1]{\mathcal{#1}}
\providecommand{\mh}[1]{\hat{#1}}

\newtheorem{definition}{Definition}

\newtheorem{conjecture}{Conjecture}

\usepackage[allcolors=RoyalBlue,colorlinks=true]{hyperref}
\usepackage{cleveref}

\title{Prospective Learning: \\ Principled Extrapolation to the Future}

\author{
    The Future Learning Collective
}

\collasfinalcopy 

\begin{document}

\maketitle

\begin{abstract}
Learning is a process which can update decision rules, based on past experience, such that future performance improves. 
Traditionally, machine learning is often evaluated under the assumption that the future will be identical to the past in distribution or change adversarially. But these assumptions can be either too optimistic or pessimistic for many problems in the real world.
Real world scenarios evolve over multiple spatiotemporal scales with  partially predictable dynamics. 
Here we reformulate the learning problem to one that centers around this idea of dynamic futures that are partially learnable.
We conjecture that certain sequences of tasks are not retrospectively learnable (in which the data distribution is fixed), but are prospectively learnable (in which distributions may be dynamic), suggesting that prospective learning is more difficult in kind than retrospective learning. 
We argue that prospective learning more accurately characterizes many real world problems that (1) currently stymie existing artificial intelligence solutions and/or (2) lack adequate explanations for how natural intelligences solve them. Thus, studying prospective learning will lead to deeper insights and solutions to currently vexing challenges in both natural and artificial intelligences.
\end{abstract}
\section{Introduction}

All learning is for the future. Learning involves updating decision rules or policies, based on past experiences, to improve \textit{future} performance. The dogma of much of machine learning is that future scenarios faced by the learner are similar to past ones, e.g., all data, including what the learner learns from and what the learner gets tested on, are independent and sampled from the same probability distribution~\citep{Glivenko1933-wm,Cantelli1933-vj, Vapnik1971-um, vapnik1991principles, Valiant1984-dx}. This is not specific to supervised learning: algorithms for sub-fields ranging from robust learning~\citep{White1982-ch, Huber2009-qu} (test distribution can be slightly different from the training distribution), online learning~\citep{Cesa-Bianchi2006-tv, Orabona2019-ux} (data distribution can arbitrarily drift over time) to adversarial learning~\citep{Madry2018-gy} (test distribution can be adversarial) and reinforcement learning~\citep{bellman1954theory,Recht2019-fs} (data distribution changes as the learner procures more data), all use variants of this assumption.
We refer to these algorithms as \textit{retrospective learners} because they learn representations that are optimal for past data, not future data. Retrospective learning is certainly useful: it is the basis for, arguably all, real-world learning systems today. Modifications to the core retrospective assumption, e.g., covariate shift detection, domain adaptation, meta-learning and continual learning, help bridge the gap that arises from using representations that are optimal on past data to make inferences on future data. Each of these ideas makes a specific assumption about future data, e.g., covariate shift assumes that only the marginal on the inputs changes, meta-learning assumes that new tasks are similar to past tasks. Learning systems can fail catastrophically when they are faced with scenarios outside these assumptions~\citep{Giurfa2001-nr, Vermaercke2014-pb, Fleuret2011-dd, Kim2018-sf, yao2022wild, lazaridou2021mind}. 

This paper attempts to rethink the fundamental assumptions of learning to address the above gap. We propose the concept of \textbf{prospective learning}, a generalization of retrospective learning, that relaxes the assumption that past (training) and future (test) distributions are identical and stationary. The goal of prospective learning explicitly emphasizes the ability to predict the future, or prospection. In comparison, adversarial learning or online
learning make no attempt at predicting the future. While the idea of prospection is by no means new~\citep{seligman2013navigating}, its role has largely been neglected in machine learning. Prospective learning embodies the eventual goal of machine learning which is to make predictions for the foreseeable future. 

\section{A failure mode of retrospective learning}

\begin{figure}[!htb]
    \begin{center}
        \includegraphics[width=\textwidth]{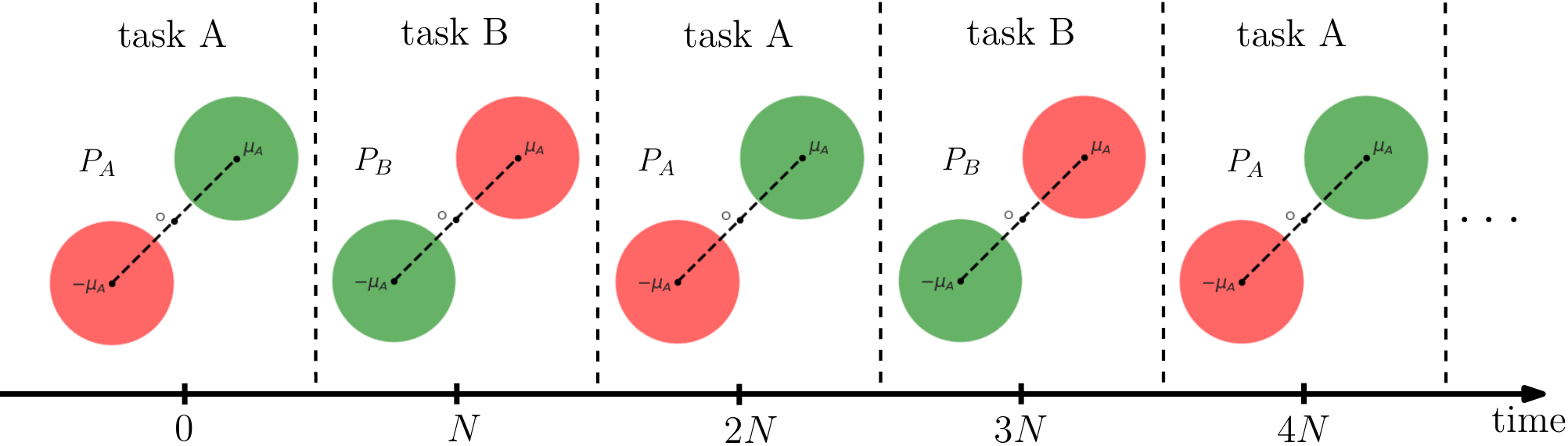}
    \end{center}
    \caption{This figure depicts an infinite sequence of data distributions formed by alternating between two distributions, $P_A$ and $P_B$. Both distributions define binary classification problems, namely task A and task B, with 2-dimensional inputs where the classes are separable using a single line. A single hypothesis, that doesn't change with time, cannot generalize arbitrarily well to this sequence of distributions. However, the sequence of distributions evolve according to a simple alternating pattern which can be exploited to generalize to tasks in the future. A prospective learner should ideally leverage this pattern by modeling the dynamics under which the tasks evolve with time.}
    \label{fig:sequence_task}

\end{figure}
 
Consider an infinite sequence of distributions denoted by $P = \{ P_t \}_{t \in \mathbb{N}}$ where time $t$ is a natural number. We see data from the first $t'$ distributions and we seek a hypothesis that generalizes to distributions that we will see in the future. 

We illustrate the insufficiency of retrospective learning using a simple example. Consider two distributions $P_A$ and $P_B$; both $P_A$ and $P_B$ define binary classification tasks with 2-dimensional inputs. The classes in both distributions can be perfectly separated by a linear classifier. Consider the sequence of distributions $P$, formed by alternating between $P_A$ and $P_B$ at regular intervals. We would like to design a learner $L$ that learns using a finite amount of data and generalizes to the infinite sequence of distributions depicted in~\Figref{fig:sequence_task}.

No single hypothesis can achieve an arbitrarily small error on both $P_A$ and $P_B$ simultaneously. 
As a result, if a retrospective learner picks a hypothesis that is unchanged over time, then it cannot achieve an arbitrarily small error on the sequence of distributions even if the learner has access to a large number of samples. 

We can also use methods from continual or meta-learning to tackle a sequence of distributions. At every time step, the learner encounters an unknown distribution and uses a few samples from this distribution in order to build a hypothesis for it. Even if the learner has seen a large number of distributions, it still requires at least 1 sample at every time step. Such a learner lowers the sample complexity of learning the unknown task with the help of a learnt inductive bias~\citep{Baxter2000-le}. However, it still fails to model one key aspect of the problem.

Many standard methods do not predict the future sequence of distributions; this is helpful if the sequence of distributions follows some pattern like in the example in~\Figref{fig:sequence_task}. For this example, if we know that the tasks alternate at every $N$ time steps, then we can maintain two different hypotheses---one for $P_A$ and another for $P_B$---and alternate between them. This helps us generalize better but more importantly allows us to do so using fewer samples. We do not need additional samples when the task switches  to adapt to an unknown distribution, since we can perfectly predict the future sequence of distributions. In general,  data distributions can change with time, and approximating the underlying dynamics of distribution shift can help us build a hypothesis that predicts the next distribution.

\section{The theoretical foundations for prospective learning}

\subsection{Retrospective learning}

To formalize prospective learning, we contrast it with traditional (retrospective) learning~\citep{Mohri2018-tf}. Consider a distribution $P$ over inputs $x$ and outputs $y$, in which case we have a supervised learning problem. A learning task is generally associated with a distribution $P$, a hypothesis class $\mc{H}$, and a risk function $R(\cdot)$. For simplicity, in this work, we represent a learning task by its associated distribution $P$, assuming that $\mc{H}$ and $R(\cdot)$ are fixed. Therefore, we use the terms ``task'' and ``distribution'' interchangeably. Using the dataset $D$, drawn independently and identically from $P \in \mc{P}$, we seek a hypothesis $h$ that minimizes the risk $R(\cdot)$. The hypothesis $h$ is selected from the hypothesis class $\mathcal{H}$. The risk of a hypothesis $h$ can be defined as $R(h) = \mathbb{E}_{(x, y) \sim P}\left[ \ell((x,y), h) \right]$ where $\ell$ is the loss function. For a classification problem, we often consider the 0-1 loss, $\ell((x,y), h) = \mathds{1} \{h(x) \neq y\}$, and for a regression problem, we often consider the mean-squared error $\ell((x,y), h) = (h(x) - y)^2$.

Classical machine learning posits that a learning task is unchanged with respect to time $t$, i.e., $ P_t \equiv P;\ \forall t$~\citep{Glivenko1933-wm, Cantelli1933-vj}. One may formalize learning for a class of problems using the probably approximately correct (PAC) model~\citep{Valiant1984-dx,Vapnik1971-um}. Under this model, we seek a learning algorithm that achieves an arbitrarily small risk on any given task $P$ using samples from that task.


\begin{definition}[PAC-learnability]
    (From \citet{Valiant1984-dx, kearns1994introduction}) A family of tasks $\mathcal{P}$ is PAC-learnable using hypothesis class $\mathcal{H}$ if there exists a learner $L$ with the following property:  For all $\epsilon, \delta > 0$ and tasks $P \in \mathcal{P}$, if $L$ has access to dataset $D$ of at least $n$ independent and identically distributed samples from $P$, then it outputs a hypothesis $\hat h \in \mathcal{H}$ which satisfies
    \begin{equation} \label{eq:RetL}
        \mathbb{P}_D[ |R(\hat h) - R(h^*)| < \epsilon] \geq 1 -\delta,
    \end{equation}
    where $\mathbb{P}_D$ is the probability over draws of the dataset $D$ and $h^*$ is the optimal hypothesis from the hypothesis class $\mathcal{H}$. The learner $L$ is said to be efficiently PAC-learnable if the sample complexity $n$ is a polynomial function of $\epsilon^{-1}$ and $\delta^{-1}$.
\end{definition}
\noindent

While~\ineqref{eq:RetL} is relatively simple, it can be used to frame the foundation of the most recent AI revolution, spanning computer vision \citep{Krizhevsky2012-er}, natural language processing \citep{Devlin2019-sx}, diagnostics \citep{McKinney2020-ft} and protein folding \citep{Senior2020-ea}. 
Yet, its limitations have been widely acknowledged for decades in ML and statistics (see~\citet{Bozinovski1976-ij, Hand2006-qc}). Specifically, real-world data rarely exhibit such simple dynamics. Specifically, the assumption that training (past) data and test/evaluation (future) data are sampled from the exact same fixed distribution is woefully inadequate.  Empirically, modern deep learning approaches continue to fail in scenarios that are trivially easy for humans \citep{Madry2018-gy, Alcorn2019-ks}, and which we believe would benefit from prospective learning capabilities. Moreover,  existing retrospective learning theory is inadequate to characterize most of the above generalized learning scenarios~\citep{Thrun1996-kk} (though see \citet{Arjovsky2021-mm,Baxter2000-le} for some compelling results). 

\subsection{Prospective learning}

Sub-fields like online~\citep{Cesa-Bianchi2006-tv}, transfer~\citep{Bozinovski1976-ij}, multitask~\citep{Bengio1992-pe, Caruana1997-mm} meta-~\citep{Baxter2000-le, Wang2020-dy}, continual~\citep{Ring1994-un}, and lifelong learning~\citep{Thrun1995-qy} present different ways to generalize the retrospective learner defined in~\ineqref{eq:RetL}. Most generalizations focus on \textit{adapting} to new tasks; given data from a set of tasks, the goal is to adapt to a new task using a small number of samples from it. On the other hand, a prospective learner attempts to model how the tasks evolve with time; the emphasis is on \textit{predicting} what new tasks will look like prospectively, as opposed to tackling them retrospectively.

We propose a working formal definition of prospective learning which generalizes the classical retrospective learning scenario described by~\ineqref{eq:RetL}. 
Instead of learning a task where the distribution, risk, and hypothesis are fixed across time, we assume that they can evolve over time.  For simplicity, hereafter we keep the risk and hypothesis class fixed, and assume that only the data distribution changes over time. To formalize this idea, we consider a fixed  sequence of distributions $P := \{ P_t \}_{t \geq 0}$. Here, $P_t$ is the distribution at time $t$. Time $t$ can take values from $[0, \infty)$ or be limited to the set of natural numbers, among other choices.

\begin{figure}
    \centering
    \includegraphics[width=0.77\textwidth]{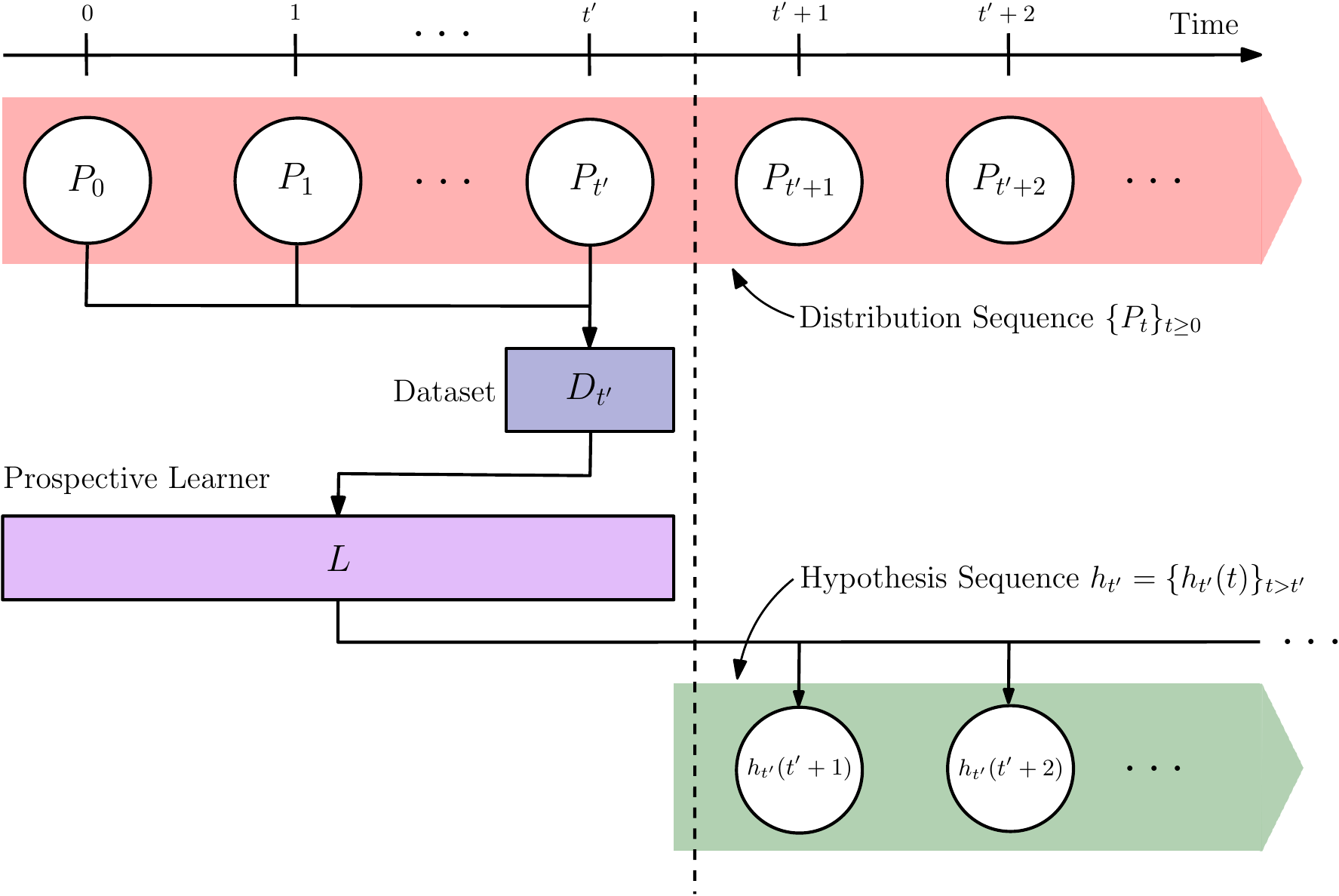}
    \caption{A schematic diagram illustrating the objective of prospective learning and the setup leading up to the~\defref{def:ProL}. $P = \{ P_t \}_{t \geq 0}$ is the sequence of distributions and $D_{t'}$ denotes a dataset drawn from $P$, over the finite time interval $0 \leq t \leq t'$. Using the prospective learner $L$ that takes $D_{t'}$ as input, we wish to estimate a time-indexed sequence of decision functions or hypotheses $\hat h_{t'}(t)$ that generalizes to future tasks encountered after time $t'$.  Therefore, the focus of prospective learning is to learn a sequence of hypotheses from data received up to $t'$ such that it generalizes to future tasks after time $t'$.}
    \label{fig:PL-diagram}
\end{figure}

Let $D_{t'}$ denote a dataset drawn from $P$ before time $t'$. Using a prospective learner $L$ that takes $D_{t'}$ as input, we wish to estimate a time-indexed sequence of decision functions or hypotheses that generalizes to future tasks encountered after time $t'$. We denote this estimated hypothesis sequence by $\hat h = \{ \hat h_t(\cdot) \}_{t \geq 0}$ where $\hat h_t$ is the hypothesis at time $t$. 
We provide a visual illustration of the objective of prospective learning in~\Figref{fig:PL-diagram}.

Finally, let $\mc{H}$ denote a hypothesis class that contains sequences of hypotheses. For a hypothesis sequence $h \in \mc{H}$, we can define the risk at time $t$, assuming we have supervised learning problems, as $R_t(h_t) = \mathbb{E}_{(x, y) \sim P_t}\left[ \ell((x, y), h_t) \right]$ where $\ell$ is some choice of a loss function. We can now provide a formal definition of prospective-learnability of the sequence $P$ using the hypothesis class $\mc{H}$ as follows. 

\begin{definition}[Prospective-Learnability] \label{def:ProL} 
Consider a sequence of distributions $P := \{ P_t \}_{t \geq 0}$ and a hypothesis class $\mc{H}$. We say that $\mc{H}$ is prospectively-learnable with respect to $P$ and a chosen reference hypothesis sequence $h$, with precision $\epsilon > 0$ and confidence $\delta > 0$ if there exists a learner $L$ and $\bar t \geq 0$ that satisfies the following property: for any $t' > \bar t$, if $L$ has access to a dataset $D_{t'}$ with at least $n_{t'}$ samples drawn from $P$ up to time $t'$, then it outputs a hypothesis sequence $\hat h \in \mc{H}$ such that it satisfies
\begin{equation}    \label{eq:ProL-condition}
 \lim_{T \to \infty} \frac{1}{T-t'} \int_{t'}^T \text{d}t\ \mathbb{P}_{D_{t'}}[ | R_t(\hat h) - R_t(h) | < \epsilon] \geq 1 -\delta.
\end{equation}
Here, $n_{t'}$ and $\bar t$ would depend on the sequence $P$, the complexity of the hypothesis class $\mc{H}$, the sequence of risks $\{R_t\}_{t \geq 0}$, and the choice of $\epsilon$ and $\delta$. 
\end{definition}

In words, the sequence of hypotheses $\hat h$, learned by $L$ from all the data points $D_{t'}$ received up to time $t' > \bar{t}$, makes predictions, with precision no worse than $\epsilon$ and with a probability of at least  $1 -\delta$ on future tasks beyond time $t'$, with respect to a reference hypothesis sequence $h$.

\defref{def:ProL} offers some flexibility when choosing the reference hypothesis sequence $h$; the performance of the prospective learner $L$ is measured against this reference. Some reasonable choices for $h$ are:

\begin{enumerate}
    \item Strong reference $h^\ast$, the optimal sequence of hypotheses given the entire distribution sequence $P$ where $h^\ast_t$ is the optimal hypothesis for $P_t$;
    \item Weak reference $h^0$, the sequence of hypotheses that performs at random chance where $h^0_t$ is the hypothesis that performs at random chance with respect to $P_{t}$. 
\end{enumerate}

Depending on the ``strictness'' of these choices, the class of sequences $P$ that are prospectively-learnable according to~\defref{def:ProL} will vary. 

In contrast to a retrospective learner, a prospective learner should respect~\ineqref{eq:ProL-condition} for all $t' > \bar t$ where $\bar t$ is finite. This is a natural condition to impose because the prospective learner needs to see a few tasks before it can build an accurate model of how the tasks evolve in the future. Here, $\bar t$ is the minimum amount of time that we need to observe data from the sequence $P$ in order to be prospectively learnable. Therefore, $\bar t$ is a function of $P$, the hypothesis class $\mc{H}$, the risk, $R$, in addition to $\epsilon, \delta$.

Equipped with this definition of prospective learnability, we can attempt to understand the properties of prospectively learnable sequences of distributions. Once we have defined the sequence of distributions that governs a given prospective learning setting (that is, those that govern changes in distributions, risks, and constraints), defining quantities that we use in retrospective learning theory, such as VC-dimension \citep{vapnik2015uniform}, is analogous. In this way, the primary research question for prospective learning is determining which distribution sequences are intrinsically low-dimensional (e.g., have a dependence structure), because otherwise learning cannot be achieved regardless of sample complexity.

When designing and implementing algorithms to perform prospective learning, it is important to determine the desirable properties of these algorithms and their output hypotheses. We state the following conjecture as an entry point to the discussion about the hypothesis sequences learned by prospective-learners.

\begin{conjecture}[Continuum Hypothesis of Learning Hypotheses] \label{conj:hypo-nature}
    \normalfont Suppose $P = \{ P_t \}_{t \geq 0}$ is a sequence of distributions. We conjecture that there are three distinct complexity classes for learning.
    \begin{enumerate}
        \item $P$ is PAC-Learnable: With the introduction of potentially time-varying $P_t$, $R_t$ and $\mc{H}_t$, PAC-learnability takes a different form.  Specifically, whenever any of the above change in such a way that the relative risk, $|R_t(\mh{h}) - R_t(h) |$ remains bounded above by $\epsilon$ with probability $1-\delta$, then it remains PAC-learnable.  A simple example of such a scenario is in a classification problem, where the posteriors can change dramatically, but the discriminant boundaries remain fixed, so the relative risk (assuming risk is expected $0-1$ loss) remains constant. 
        \item $P$ is prospectively learnable: 
        This is where the task evolves  such that some complexity measure of the sequence $P$ is low, in which case, we conjecture that a prospective learner can learn a sequence of hypotheses $\hat h \in \mc{H}$ after observing $n_{t'}$ amount of data drawn from $P$ for a duration $t' > \bar t$ such that $\hat h$ can make accurate predictions in the future. Note that $\mc{H}, \bar t$ and $n_{t'}$ would depend on the complexity of $P$. For instance, when the complexity of $P$ is high, we may need to search for $\hat h$ from a more complex class of hypotheses and observe a larger amount of data from $P$ for a longer time. 
        \item $P$ is not prospectively learnable:  Learning such a sequence of distributions would be impossible and it requires us to observe the sequence $P$ for an infinite period of time or collect an infinite amount of data from it. We deem such sequences too complicated to learn prospectively.
    \end{enumerate}
    \end{conjecture}

While we do not propose a complexity measure for a distribution sequence in this work, we note that quantities such as predictive information \citep{bialek2001predictability}, which compute the mutual information between the past and the future of a time series, can inspire the construction of an appropriate measure. We leave this and the discovery of further desirable properties of hypothesis sequences suitable for prospective learning as future work. 

\conref{conj:hypo-nature} motivates the development of prospective learning algorithms that can output dynamic sequences of hypotheses. Such learners must  update  hypotheses as time evolves. Note that prospective learning as defined here is equivalent to retrospective learning if the distribution sequence is a constant. Thus traditional retrospective learning can be seen as a special case of prospective learning. But this formulation of prospective learning is also quite general, and it has relations with continual-, lifelong-, transfer-, meta-, online-, and OOD-learning. However, we leave the task of verifying whether the learnability of these paradigms is encapsulated by \defref{def:ProL} as future work.

\section{Related Work}~\label{s:related}



\paragraph{Online Learning} Both online learning \citep{Cesa-Bianchi2006-tv, hoi2021online, ying2008online} and prospective learning deal with task sequences.
The goal of the online learner is to iteratively estimate the hypothesis such that it minimizes regret, which is the difference between the learner’s risk and the risk of the best model in hindsight. It is usually studied using a worst-case analysis and we assume that the sequence of tasks can even be generated by an adversary. Many algorithms use a follow-the-leader (FTL) strategy which selects the best model averaged over the past, since it achieves an asymptotically optimal regret. As a result, an online learner that uses FTL doesn't consider the temporal dynamics of the sequence and has no prospective abilities. The same is true for other popular algorithms like online gradient descent (OGD) and its variants.

\begin{figure}[h]
    \centering
    \begin{subfigure}[b]{0.49\linewidth}
        \centering
        \includegraphics[width=0.9\textwidth]{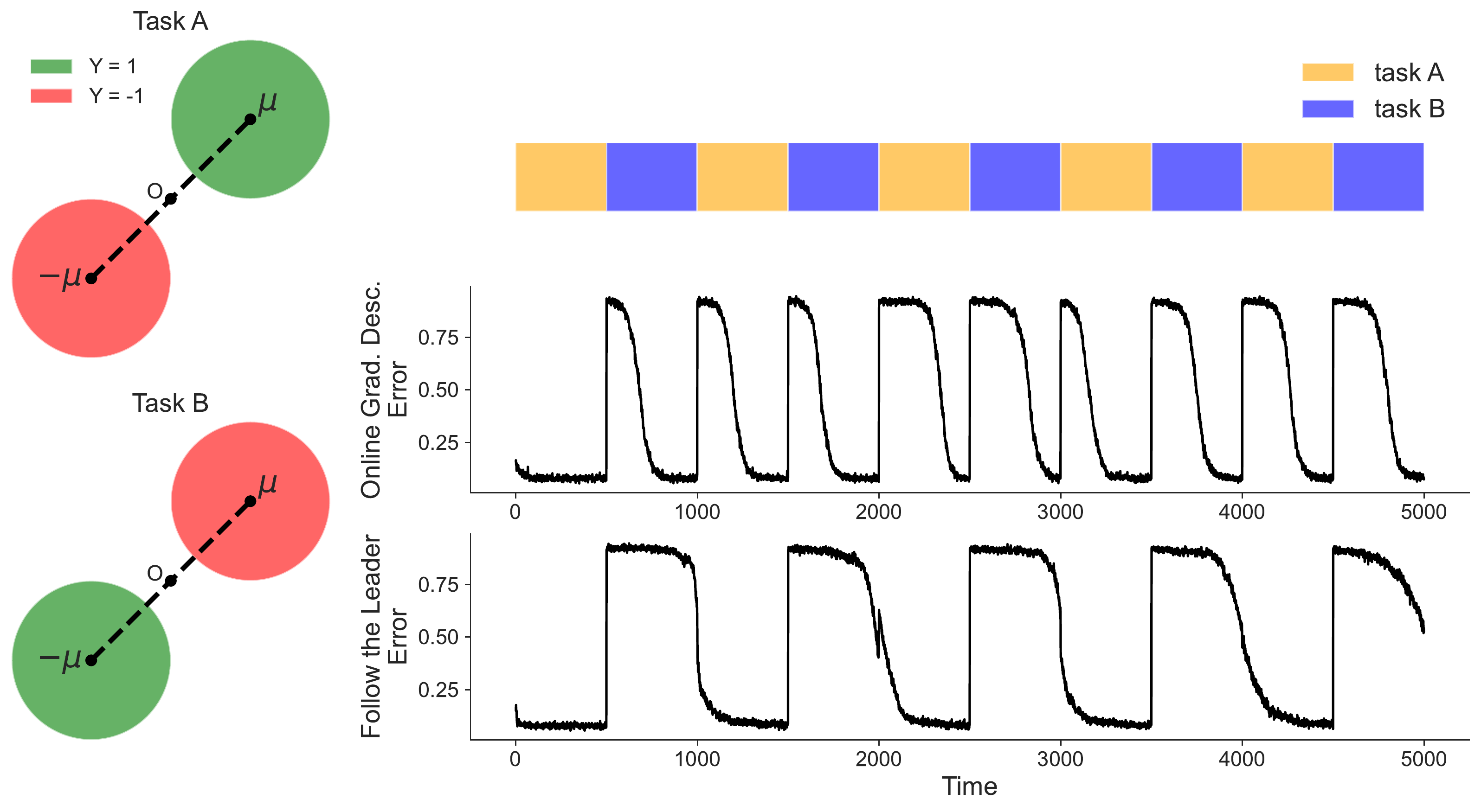}
        \caption{}
        \label{fig:online_seq1}
    \end{subfigure}
    \begin{subfigure}[b]{0.49\linewidth}
        \centering
        \includegraphics[width=0.9\textwidth]{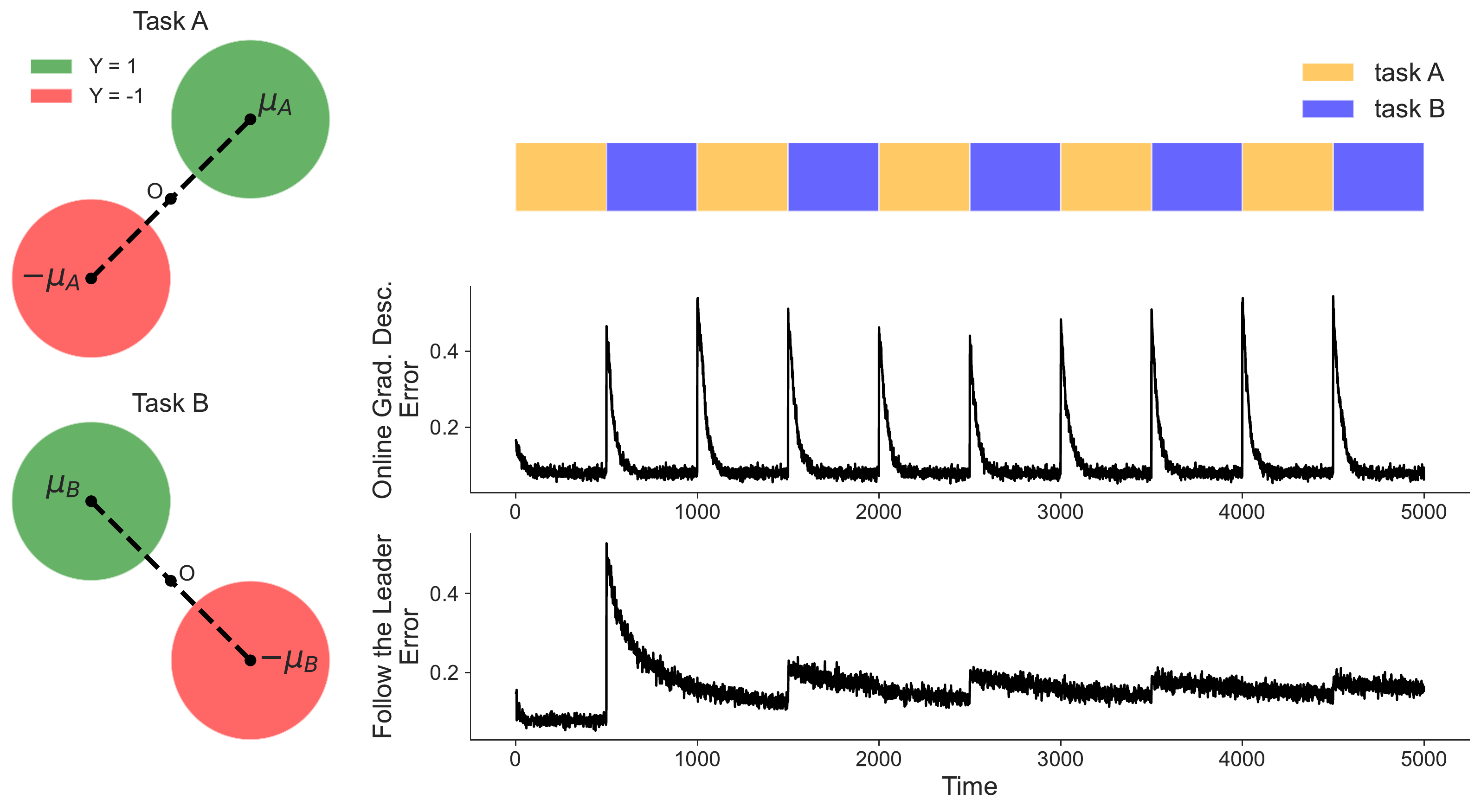}
        \caption{}
        \label{fig:online_seq2}
    \end{subfigure}
    \caption{We consider two different alternating sequences of tasks where the learning task shifts between task A and task B periodically every 500 time-steps. In this online setting, we see one new sample every time-step. We plot the test error of the current task against the number of time-steps for FTL and OGD.\\[0.1em]
    \textbf{(a)} For task A, the samples from one class belong to a Gaussian with mean $[1, 1]^\top$ and samples from the other class belong to a Gaussian with mean $[-1, -1]^\top$. Both Gaussians have isotropic covariance $I$. Task B is similar to Task A, except that the distributions for the two labels are interchanged. The OGD algorithm's error peaks at each task switch and slowly reduces afterwards as it learns the new task by unlearning the one before. On the other hand, the FTL algorithm seems to learn task A successfully, but completely fails at learning task B. We speculate that for FTL, the model weights are biased to task A.\\[0.1em]
    \textbf{(b)} For task A, the samples from one class belong to a Gaussian with mean $[1, 1]^\top$ and samples from the other class belong to a Gaussian with mean $[-1, -1]^\top$. For task B, the samples from one class belong to a Gaussian with mean $[1, -1]^\top$ and samples from the other class belong to a Gaussian with mean $[-1, 1]^\top$. All Gaussians have isotropic covariance $I$. Unlike (a), the class conditional means are completely different making the task sequence slightly easier. 
    OGD learns each task much faster, but test error of the current task peaks at each task switch. Interestingly, FTL is able to progressively reduce its error as it observes more data from the sequence. It manages to converge to a fixed hypothesis that acheives reasonably low error on both task A and task B. However, FTL does not converge to the Bayes optimal hypothesis for either task since that requires it to be prospective. \\[0.1em]
    In conclusion, neither OGD nor FTL attempt to model the temporal dynamics of the task sequence. We expect a (reasonable) prospective learner would achieve smaller error on both problems.}
    \label{fig:online-seq}
\end{figure}

In~\Figref{fig:online-seq}, we evaluate FTL and OGD on two different alternating sequence of binary classification tasks in the online setting. In~\Figref{fig:online_seq1}, we consider two tasks with their labels flipped and observe that both OGD and FTL usually have high test error when the task switches. Hence both learners fail to exploit the dynamics of the task sequence. In~\Figref{fig:online_seq2}, we consider an alternating sequence of tasks where the first and second task have different class-conditional means. We observe that this sequence is weakly prospective learnable by FTL since it converges to some fixed hypothesis that works better than random on both tasks. We expect that a \textit{bona fide} prospective learner would successfully and optimally tackle both of these problems as it would model the temporal dynamics of the task sequence.

\paragraph{Online Meta-Learning} Online meta-learning \citep{finn2019online, denevi2019online, yao2020online} refers to a class of algorithms that implement meta-learning updates in an online manner. These methods seek to minimize a variant of regret which inspires the use of the follow-the-meta-leader strategy%
~\citep{finn2019online}. In practice, this  translates to implementing model-agnostic meta-learning (MAML)~\citep{finn2017model} but with the tasks sampled uniformly at random from all previously seen tasks. While this minimizes the worst-case regret, this choice is sub-optimal if we know that the task sequence evolves with time according to a predictable pattern. Hence, methods like online-MAML will fail to be strongly prospective learnable for some task sequences.

\paragraph{Continual Learning} Unlike PAC Learning and online learning, continual learning \citep{van2019three, zenke2017continual, ramesh2021model, vogelstein2020representation, geisa2021towards} lacks a universally agreed-upon definition. In the current literature, it often refers to a wide gamut of scenarios that vary in the goal (e.g., avoiding catastrophic forgetting, positive forward/backward transfer), or storage and computational constraints. 
A lack of a universally agreed-upon definition limits our ability to formally delineate differences.  That said, prospective learning resembles continual learning, albeit with some key differences. Suppose we would like to tackle a sequence of tasks where the task switches from A to B, and back, at regular intervals. Many existing continual learning methods will eventually learn both task A and B, and will therefore be able to switch back and forth with high accuracy---but  shortly after the task switches. In other words, current continual algorithms will \textbf{fail to predict when the task will switch}. In contrast, a prospective learner, that satisfies \defref{def:ProL}, will learn how the sequence of tasks evolves, and therefore predict when the task switches. Formally, this learner should perform at least as well as the continual learner that does not model the dynamics of the tasks. In simple terms, a continual learner would iteratively ``adapt'' to future tasks while a prospective learner would ``anticipate'' them \textit{a priori}. We certainly expect continual learners to benefit from prospective abilities and believe learners should ideally incorporate both adaptation and anticipation to actively learn from and make inferences about the future.

\paragraph{Streaming Learning} The setting of streaming learning is also closely related to prospective learning but the objective can often vary. For example, the goal of \citet{gama2013evaluating} is to minimize the error on samples from an underlying stationary process. More recent works on streaming learning such as \citet{hayes2020remind} minimize the error on a fixed held-out dataset or on all the past data---neither of which emphasizes prospection.

\paragraph{Out-of-distribution (OOD) Generalization} The OOD generalization problem \citep{shen2021towards, geisa2021towards, arjovsky2020out, ye2022ood} seeks to minimize the generalization error on an OOD task where the hypothesis is selected using data from a potentially different set of domains. OOD is typically defined for batch data, not sequential data. That said, we recover the OOD generalization setting if we assume that the distribution shifts at time $t’$---which is the point up to which we see data--- and then never changes again.


\paragraph{Few-shot Learning} Few-shot learning \citep{triantafillou2019meta, wang2020generalizing} refers to a broad range of settings where the model adapts to a new task using limited amounts of labeled data. Usually, the adaptation step has a limited computational budget but this is difficult to write down formally. Prospective learners are also expected to learn using very few samples with one key difference: they generalize to future tasks using ‘time’ as another context variable.



It is evident that when the stochastic process is a constant process, prospective learning becomes equivalent to PAC learning. Thus PAC learning can be seen as a special case of prospective learning. Furthermore, as discussed above, prospective learning is quite general and it has relations with various learning paradigms such as continual-, lifelong-, transfer-, meta-, online-, and OOD-learning. However, learning paradigms besides PAC learning and online learning lack formal and universally agreed-upon definitions of learnability. Therefore, concrete comparison of these paradigms with prospective learning is limited.







\section{Discussion}
\label{sec:future}

Classical machine learning 
limits the ability of learning systems to effectively learn in environments that (somewhat) predictably change. Here we have proposed an alternative to the classical learning problem that works around the  assumption  that training (past) and test (future) distributions are not identical, but predictably related to some degree. This reframing of the learning problem can produce flexible agents that effectively adapt in novel contexts with invariance along key dimensions relevant to learning. As a result, algorithms that follow a prospective learning framework can effectively learn for the future, overcoming a fundamental limitation of most modern AI.


We have provided an answer to the question ``What is prospective learning?'', yet we have largely ignored the broader question of ``how is prospective learning implemented?''. Assuming that natural intelligences (NI; including, but not necessarily limited to, humans) implement prospective learning~\cite{seligman2013navigating}, motivates synergistic research between NI and AI. 
Carefully designed, ecologically appropriate experiments across species, phyla, and substrates, will enable (i) quantifying the limitations and capabilities of existing NIs with respect to prospective learning criteria, and (ii) refining the mechanistic hypotheses generated by the AI theory. 

Discovering \textit{how} prospective learning is implemented in NIs may accelerate incorporating prospective learning capabilities in AI. For example, experiments in NI across taxa can establish milestones for experiments in AI (e.g., ~\citep{Turing1950-or}). Implementing and deploying AIs and observing NIs exhibiting prospective learning `in the wild' will provide real-world evidence to deepen our understanding of learning for the future. These AI implementations could be purely software,   leverage specialized (neuromorphic) hardware~\citep{Vogelstein2007-wz,Davies2019-jr}, or even include wetware and hybrids~\citep{Silva2020-bz}, as prospective learning is, by definition, substrate independent.

In closing, there are many ways in which how we frame intelligence in machine learning differs from the kinds of problems that we really want to (and evolved to) solve in the real world, including both objective functions and prior knowledge. Other discrepancies  are hard to even formalize~\citep{Mitchell2019-hp}. Here we have argued that the problem of  shifts in the world can be formulated considerably better than to just appeal to out of distribution learning. We highlight how many current developments in machine learning can be seen as steps towards prospective learning. We believe that formulating the learning problem as a prospective one will produce important synergies between multiple branches of machine learning and cognitive sciences, providing a more thorough foundation and new ways of providing logical precision.

\section*{Acknowledgements}

This paper was supported by an NSF AI Institute Planning award (\#2020312), NSF-Simons Research Collaborations on the Mathematical and Scientific Foundations of Deep Learning (MoDL), NSF-Simons Research Collaborations on Transferable, Hierarchical, Expressive, Optimal, Robust, Interpretable Networks (THEORINET, \#2031985), Microsoft Research, and DARPA. RR and PC were supported by grants from the National Science Foundation (IIS-2145164, CCF-2212519), the Office of Naval Research (N00014-22-1-2255), and cloud computing credits from Amazon Web Services. The authors would like to especially thank Kathryn Vogelstein for putting up with endless meetings at the Vogelstein residence in order to make these ideas come to life.

\newcommand{\Yinyang}[1][1]{%
    \begin{tikzpicture}[scale=#1*0.07]
      \draw[line width = #1*0.05mm,transform canvas={yshift=0.02cm}] (0,0) circle (1cm);
      \path[fill=black,transform canvas={yshift=0.02cm}] (90:1cm) arc (90:-90:0.5cm)
                        (0,0)    arc (90:270:0.5cm)
                        (0,-1cm) arc (-90:-270:1cm);
    \end{tikzpicture}}

\section*{The Future Learning Collective}

Ashwin De Silva$^{1*}$;
Rahul Ramesh$^{2*}$;
Lyle Ungar$^{2}$;
Marshall Hussain Shuler$^{1}$;
Noah J. Cowan$^{1}$;
Michael Platt$^{2}$;
Chen Li$^{1}$;
{Leyla Isik}$^{1}$;
Seung-Eon Roh$^{1}$;
Adam Charles$^{1}$;
Archana Venkataraman$^{1}$;
Brian Caffo$^{1}$;
Javier J. How$^{1}$;
Justus M Kebschull$^{1}$;
John W. Krakauer$^{1}$;
Maxim Bichuch$^{1}$;
Kaleab Alemayehu Kinfu$^{1}$;
Eva Yezerets$^{1}$;
Dinesh Jayaraman$^{2}$;
Jong M. Shin$^{1}$;
Soledad Villar$^{1}$;
Ian Phillips$^{1}$;
Carey E. Priebe$^{1}$;
Thomas Hartung$^{1}$;
Michael I. Miller$^{1}$;
Jayanta Dey$^{1}$;
Ningyuan (Teresa) Huang$^{1}$;
Eric Eaton$^{2}$;
Ralph Etienne-Cummings$^{1}$;
Elizabeth L. Ogburn$^{1}$;
Randal Burns$^{1}$;
Onyema Osuagwu$^{4}$;
Brett Mensh$^{5}$;
Alysson R. Muotri$^{6}$;
Julia Brown$^{7}$;
Chris White$^{8}$;
Weiwei Yang$^{8}$;
Andrei A. Rusu$^{9}$
{Timothy Verstynen}$^{3}$;
{Konrad P. Kording}$^{2}$; 
Pratik Chaudhari$^{2}$\textsuperscript{\Letter};
{Joshua T. Vogelstein}$^{1}$\textsuperscript{\Letter};

%
%

\ 

\noindent
$^{1}$ Johns Hopkins University; 
$^{2}$ University of Pennsylvania;
$^{3}$ Carnegie Mellon University;
$^{4}$ Morgan State University
$^{5}$ Howard Hughes Medical Institute;
$^{6}$ University of California, San Diego
$^{7}$ MindX;
$^{8}$ Microsoft Research
$^{9}$ DeepMind

\noindent 
$^*$ Equal contribution \\ 
\textsuperscript{\Letter} Corresponding authors:
\href{mailto:pratikac@seas.upenn.edu}{pratikac@seas.upenn.edu}, \href{mailto:jovo@progl.ai}{jovo@progl.ai}

\clearpage
\bibliography{backmatter/references.bib}

\begin{thebibliography}{61}
\providecommand{\natexlab}[1]{#1}
\providecommand{\url}[1]{\texttt{#1}}
\expandafter\ifx\csname urlstyle\endcsname\relax
  \providecommand{\doi}[1]{doi: #1}\else
  \providecommand{\doi}{doi: \begingroup \urlstyle{rm}\Url}\fi

\bibitem[Alcorn et~al.(2019)Alcorn, Li, Gong, Wang, Mai, Ku, and
  Nguyen]{Alcorn2019-ks}
Michael~A Alcorn, Qi~Li, Zhitao Gong, Chengfei Wang, Long Mai, Wei-Shinn Ku,
  and Anh Nguyen.
\newblock {Strike (with) a pose: Neural networks are easily fooled by strange
  poses of familiar objects}.
\newblock In \emph{{Proceedings of the IEEE/CVF Conference on Computer Vision
  and Pattern Recognition}}, pp.\  4845--4854. openaccess.thecvf.com, 2019.

\bibitem[Arjovsky(2020)]{arjovsky2020out}
Martin Arjovsky.
\newblock \emph{Out of distribution generalization in machine learning}.
\newblock PhD thesis, New York University, 2020.

\bibitem[Arjovsky(2021)]{Arjovsky2021-mm}
Martin Arjovsky.
\newblock Out of distribution generalization in machine learning, March 2021.

\bibitem[Baxter(2000)]{Baxter2000-le}
Jonathan Baxter.
\newblock A model of inductive bias learning.
\newblock \emph{J. Artif. Intell. Res.}, 12\penalty0 (1):\penalty0 149--198,
  March 2000.

\bibitem[Bellman(1954)]{bellman1954theory}
Richard Bellman.
\newblock The theory of dynamic programming.
\newblock \emph{Bulletin of the American Mathematical Society}, 60\penalty0
  (6):\penalty0 503--515, 1954.

\bibitem[Bengio et~al.(1992)Bengio, Bengio, Cloutier, and
  Gecsei]{Bengio1992-pe}
Samy Bengio, Yoshua Bengio, Jocelyn Cloutier, and Jan Gecsei.
\newblock On the optimization of a synaptic learning rule.
\newblock In \emph{Preprints Conf. Optimality in Artificial and Biological
  Neural Networks}, volume~2. researchgate.net, 1992.

\bibitem[Bialek et~al.(2001)Bialek, Nemenman, and
  Tishby]{bialek2001predictability}
William Bialek, Ilya Nemenman, and Naftali Tishby.
\newblock Predictability, complexity, and learning.
\newblock \emph{Neural computation}, 13\penalty0 (11):\penalty0 2409--2463,
  2001.

\bibitem[Bozinovski \& Fulgosi(1976)Bozinovski and Fulgosi]{Bozinovski1976-ij}
Stevo Bozinovski and Ante Fulgosi.
\newblock The influence of pattern similarity and transfer learning upon the
  training of a base perceptron {B2}.
\newblock \emph{Proceedings of Symposium Informatica}, pp.\  3--121--5, 1976.

\bibitem[Cantelli(1933)]{Cantelli1933-vj}
Francesco~Paolo Cantelli.
\newblock Sulla determinazione empirica delle leggi di probabilita.
\newblock \emph{Giorn. Ist. Ital. Attuari}, 4\penalty0 (421-424), 1933.

\bibitem[Caruana(1997)]{Caruana1997-mm}
Rich Caruana.
\newblock Multitask learning.
\newblock \emph{Mach. Learn.}, 28\penalty0 (1):\penalty0 41--75, July 1997.

\bibitem[Cesa-Bianchi \& Lugosi(2006)Cesa-Bianchi and
  Lugosi]{Cesa-Bianchi2006-tv}
Nicolo Cesa-Bianchi and Gabor Lugosi.
\newblock \emph{Prediction, Learning, and Games}.
\newblock Cambridge University Press, March 2006.

\bibitem[Davies(2019)]{Davies2019-jr}
M~Davies.
\newblock {Progress in Neuromorphic Computing : Drawing Inspiration from Nature
  for Gains in AI and Computing}.
\newblock In \emph{{2019 International Symposium on VLSI Design, Automation and
  Test (VLSI-DAT)}}, pp.\  1--1, April 2019.

\bibitem[Denevi et~al.(2019)Denevi, Stamos, Ciliberto, and
  Pontil]{denevi2019online}
Giulia Denevi, Dimitris Stamos, Carlo Ciliberto, and Massimiliano Pontil.
\newblock Online-within-online meta-learning.
\newblock \emph{Advances in Neural Information Processing Systems}, 32, 2019.

\bibitem[Devlin et~al.(2019)Devlin, Chang, Lee, and Toutanova]{Devlin2019-sx}
Jacob Devlin, Ming-Wei Chang, Kenton Lee, and Kristina Toutanova.
\newblock {BERT: Pre-training of Deep Bidirectional Transformers for Language
  Understanding}.
\newblock In \emph{{NAACL-HLT}}, January 2019.

\bibitem[Finn et~al.(2017)Finn, Abbeel, and Levine]{finn2017model}
Chelsea Finn, Pieter Abbeel, and Sergey Levine.
\newblock Model-agnostic meta-learning for fast adaptation of deep networks.
\newblock In \emph{International conference on machine learning}, pp.\
  1126--1135. PMLR, 2017.

\bibitem[Finn et~al.(2019)Finn, Rajeswaran, Kakade, and Levine]{finn2019online}
Chelsea Finn, Aravind Rajeswaran, Sham Kakade, and Sergey Levine.
\newblock Online meta-learning.
\newblock In \emph{International Conference on Machine Learning}, pp.\
  1920--1930. PMLR, 2019.

\bibitem[Fleuret et~al.(2011)Fleuret, Li, Dubout, Wampler, Yantis, and
  Geman]{Fleuret2011-dd}
Fran{\c c}ois Fleuret, Ting Li, Charles Dubout, Emma~K Wampler, Steven Yantis,
  and Donald Geman.
\newblock {Comparing machines and humans on a visual categorization test}.
\newblock \emph{Proc. Natl. Acad. Sci. U. S. A.}, 108\penalty0 (43):\penalty0
  17621--17625, October 2011.

\bibitem[Gama et~al.(2013)Gama, Sebastiao, and Rodrigues]{gama2013evaluating}
Joao Gama, Raquel Sebastiao, and Pedro~Pereira Rodrigues.
\newblock On evaluating stream learning algorithms.
\newblock \emph{Machine learning}, 90:\penalty0 317--346, 2013.

\bibitem[Geisa et~al.(2021)Geisa, Mehta, Helm, Dey, Eaton, Dick, Priebe, and
  Vogelstein]{geisa2021towards}
Ali Geisa, Ronak Mehta, Hayden~S Helm, Jayanta Dey, Eric Eaton, Jeffery Dick,
  Carey~E Priebe, and Joshua~T Vogelstein.
\newblock Towards a theory of out-of-distribution learning.
\newblock \emph{arXiv preprint arXiv:2109.14501}, 2021.

\bibitem[Giurfa et~al.(2001)Giurfa, Zhang, Jenett, Menzel, and
  Srinivasan]{Giurfa2001-nr}
M~Giurfa, S~Zhang, A~Jenett, R~Menzel, and M~V Srinivasan.
\newblock {The concepts of 'sameness' and 'difference' in an insect}.
\newblock \emph{Nature}, 410\penalty0 (6831):\penalty0 930--933, April 2001.

\bibitem[Glivenko(1933)]{Glivenko1933-wm}
V~Glivenko.
\newblock Sulla determinazione empirica delle leggi di probabilita.
\newblock \emph{Gion. Ist. Ital. Attauri.}, 4:\penalty0 92--99, 1933.

\bibitem[Hand(2006)]{Hand2006-qc}
David~J Hand.
\newblock Classifier technology and the illusion of progress.
\newblock \emph{Statistical Science}, 21\penalty0 (1):\penalty0 1--14, February
  2006.

\bibitem[Hayes et~al.(2020)Hayes, Kafle, Shrestha, Acharya, and
  Kanan]{hayes2020remind}
Tyler~L Hayes, Kushal Kafle, Robik Shrestha, Manoj Acharya, and Christopher
  Kanan.
\newblock Remind your neural network to prevent catastrophic forgetting.
\newblock In \emph{Computer Vision--ECCV 2020: 16th European Conference,
  Glasgow, UK, August 23--28, 2020, Proceedings, Part VIII 16}, pp.\  466--483.
  Springer, 2020.

\bibitem[Hoi et~al.(2021)Hoi, Sahoo, Lu, and Zhao]{hoi2021online}
Steven~CH Hoi, Doyen Sahoo, Jing Lu, and Peilin Zhao.
\newblock Online learning: A comprehensive survey.
\newblock \emph{Neurocomputing}, 459:\penalty0 249--289, 2021.

\bibitem[Huber \& Ronchetti(2009)Huber and Ronchetti]{Huber2009-qu}
Peter~J Huber and Elvezio~M Ronchetti.
\newblock \emph{{Robust Statistics}}.
\newblock Wiley, 2 edition edition, February 2009.

\bibitem[Kearns \& Vazirani(1994)Kearns and Vazirani]{kearns1994introduction}
Michael~J Kearns and Umesh Vazirani.
\newblock \emph{An introduction to computational learning theory}.
\newblock MIT press, 1994.

\bibitem[Kim et~al.(2018)Kim, Ricci, and Serre]{Kim2018-sf}
Junkyung Kim, Matthew Ricci, and Thomas Serre.
\newblock {Not-So-CLEVR: learning same-different relations strains feedforward
  neural networks}.
\newblock \emph{Interface Focus}, 8\penalty0 (4):\penalty0 20180011, August
  2018.

\bibitem[Krizhevsky et~al.(2012)Krizhevsky, Sutskever, and
  Hinton]{Krizhevsky2012-er}
A~Krizhevsky, I~Sutskever, and G~E Hinton.
\newblock Imagenet classification with deep convolutional neural networks.
\newblock \emph{Adv. Neural Inf. Process. Syst.}, 2012.

\bibitem[Lazaridou et~al.(2021)Lazaridou, Kuncoro, Gribovskaya, Agrawal, Liska,
  Terzi, Gimenez, de~Masson~d'Autume, Kocisky, Ruder,
  et~al.]{lazaridou2021mind}
Angeliki Lazaridou, Adhi Kuncoro, Elena Gribovskaya, Devang Agrawal, Adam
  Liska, Tayfun Terzi, Mai Gimenez, Cyprien de~Masson~d'Autume, Tomas Kocisky,
  Sebastian Ruder, et~al.
\newblock Mind the gap: Assessing temporal generalization in neural language
  models.
\newblock \emph{Advances in Neural Information Processing Systems},
  34:\penalty0 29348--29363, 2021.

\bibitem[Madry et~al.(2017)Madry, Makelov, Schmidt, Tsipras, and
  Vladu]{Madry2018-gy}
Aleksander Madry, Aleksandar Makelov, Ludwig Schmidt, Dimitris Tsipras, and
  Adrian Vladu.
\newblock Towards deep learning models resistant to adversarial attacks.
\newblock \emph{arXiv preprint arXiv:1706.06083}, 2017.

\bibitem[McKinney et~al.(2020)McKinney, Sieniek, Godbole, Godwin, Antropova,
  Ashrafian, Back, Chesus, Corrado, Darzi, et~al.]{McKinney2020-ft}
Scott~Mayer McKinney, Marcin Sieniek, Varun Godbole, Jonathan Godwin, Natasha
  Antropova, Hutan Ashrafian, Trevor Back, Mary Chesus, Greg~S Corrado, Ara
  Darzi, et~al.
\newblock International evaluation of an ai system for breast cancer screening.
\newblock \emph{Nature}, 577\penalty0 (7788):\penalty0 89--94, 2020.

\bibitem[Mitchell(2019)]{Mitchell2019-hp}
M~Mitchell.
\newblock {Artificial intelligence hits the barrier of meaning}.
\newblock \emph{Information}, 2019.

\bibitem[Mohri et~al.(2018)Mohri, Rostamizadeh, and Talwalkar]{Mohri2018-tf}
Mehryar Mohri, Afshin Rostamizadeh, and Ameet Talwalkar.
\newblock \emph{Foundations of Machine Learning}.
\newblock MIT Press, November 2018.

\bibitem[Orabona(2019)]{Orabona2019-ux}
Francesco Orabona.
\newblock \emph{{A Modern Introduction to Online Learning}}.
\newblock December 2019.

\bibitem[Ramesh \& Chaudhari(2021)Ramesh and Chaudhari]{ramesh2021model}
Rahul Ramesh and Pratik Chaudhari.
\newblock Model zoo: A growing" brain" that learns continually.
\newblock \emph{arXiv preprint arXiv:2106.03027}, 2021.

\bibitem[Recht(2019)]{Recht2019-fs}
Benjamin Recht.
\newblock {A Tour of Reinforcement Learning: The View from Continuous Control}.
\newblock \emph{Annual Review of Control, Robotics, and Autonomous Systems},
  May 2019.

\bibitem[Ring(1994)]{Ring1994-un}
M~B Ring.
\newblock \emph{Continual learning in reinforcement environments}.
\newblock PhD thesis, University of Texas at Austin, 1994.

\bibitem[Seligman et~al.(2013)Seligman, Railton, Baumeister, and
  Sripada]{seligman2013navigating}
Martin~EP Seligman, Peter Railton, Roy~F Baumeister, and Chandra Sripada.
\newblock Navigating into the future or driven by the past.
\newblock \emph{Perspectives on psychological science}, 8\penalty0
  (2):\penalty0 119--141, 2013.

\bibitem[Senior et~al.(2020)Senior, Evans, Jumper, Kirkpatrick, Sifre, Green,
  Qin, {\v{Z}}{\'\i}dek, Nelson, Bridgland, et~al.]{Senior2020-ea}
Andrew~W Senior, Richard Evans, John Jumper, James Kirkpatrick, Laurent Sifre,
  Tim Green, Chongli Qin, Augustin {\v{Z}}{\'\i}dek, Alexander~WR Nelson, Alex
  Bridgland, et~al.
\newblock Improved protein structure prediction using potentials from deep
  learning.
\newblock \emph{Nature}, 577\penalty0 (7792):\penalty0 706--710, 2020.

\bibitem[Shen et~al.(2021)Shen, Liu, He, Zhang, Xu, Yu, and
  Cui]{shen2021towards}
Zheyan Shen, Jiashuo Liu, Yue He, Xingxuan Zhang, Renzhe Xu, Han Yu, and Peng
  Cui.
\newblock Towards out-of-distribution generalization: A survey.
\newblock \emph{arXiv preprint arXiv:2108.13624}, 2021.

\bibitem[Silva et~al.(2020)Silva, Muotri, and White]{Silva2020-bz}
G~A Silva, A~R Muotri, and C~White.
\newblock Understanding the human brain using brain organoids and a
  {Structure-Function} theory, 2020.

\bibitem[Thrun(1996)]{Thrun1996-kk}
Sebastian Thrun.
\newblock Is learning the n-th thing any easier than learning the first?
\newblock In D~S Touretzky, M~C Mozer, and M~E Hasselmo (eds.), \emph{Advances
  in Neural Information Processing Systems 8}, pp.\  640--646. MIT Press, 1996.

\bibitem[Thrun \& Mitchell(1995)Thrun and Mitchell]{Thrun1995-qy}
Sebastian Thrun and Tom~M Mitchell.
\newblock Lifelong robot learning.
\newblock \emph{Rob. Auton. Syst.}, 15\penalty0 (1):\penalty0 25--46, July
  1995.

\bibitem[Triantafillou et~al.(2019)Triantafillou, Zhu, Dumoulin, Lamblin, Evci,
  Xu, Goroshin, Gelada, Swersky, Manzagol, et~al.]{triantafillou2019meta}
Eleni Triantafillou, Tyler Zhu, Vincent Dumoulin, Pascal Lamblin, Utku Evci,
  Kelvin Xu, Ross Goroshin, Carles Gelada, Kevin Swersky, Pierre-Antoine
  Manzagol, et~al.
\newblock Meta-dataset: A dataset of datasets for learning to learn from few
  examples.
\newblock \emph{arXiv preprint arXiv:1903.03096}, 2019.

\bibitem[Turing(1950)]{Turing1950-or}
A~M Turing.
\newblock {COMPUTING MACHINERY AND INTELLIGENCE}.
\newblock \emph{Mind}, LIX\penalty0 (236):\penalty0 433--460, October 1950.

\bibitem[Valiant(1984)]{Valiant1984-dx}
L~G Valiant.
\newblock A theory of the learnable.
\newblock \emph{Commun. ACM}, 27\penalty0 (11):\penalty0 1134--1142, November
  1984.

\bibitem[Van~de Ven \& Tolias(2019)Van~de Ven and Tolias]{van2019three}
Gido~M Van~de Ven and Andreas~S Tolias.
\newblock Three scenarios for continual learning.
\newblock \emph{arXiv preprint arXiv:1904.07734}, 2019.

\bibitem[Vapnik \& Chervonenkis(1971)Vapnik and Chervonenkis]{Vapnik1971-um}
V~Vapnik and A~Chervonenkis.
\newblock On the uniform convergence of relative frequencies of events to their
  probabilities.
\newblock \emph{Theory Probab. Appl.}, 16\penalty0 (2):\penalty0 264--280,
  January 1971.

\bibitem[Vapnik(1991)]{vapnik1991principles}
Vladimir Vapnik.
\newblock Principles of risk minimization for learning theory.
\newblock \emph{Advances in neural information processing systems}, 4, 1991.

\bibitem[Vapnik \& Chervonenkis(2015)Vapnik and
  Chervonenkis]{vapnik2015uniform}
Vladimir~N Vapnik and A~Ya Chervonenkis.
\newblock On the uniform convergence of relative frequencies of events to their
  probabilities.
\newblock In \emph{Measures of complexity}, pp.\  11--30. Springer, 2015.

\bibitem[Vermaercke et~al.(2014)Vermaercke, Cop, Willems, D'Hooge, and Op~de
  Beeck]{Vermaercke2014-pb}
Ben Vermaercke, Elsy Cop, Sam Willems, Rudi D'Hooge, and Hans~P Op~de Beeck.
\newblock {More complex brains are not always better: rats outperform humans in
  implicit category-based generalization by implementing a similarity-based
  strategy}.
\newblock \emph{Psychon. Bull. Rev.}, 21\penalty0 (4):\penalty0 1080--1086,
  August 2014.

\bibitem[Vogelstein et~al.(2020)Vogelstein, Dey, Helm, LeVine, Mehta, Tomita,
  Xu, Geisa, Wang, van~de Ven, et~al.]{vogelstein2020representation}
Joshua~T Vogelstein, Jayanta Dey, Hayden~S Helm, Will LeVine, Ronak~D Mehta,
  Tyler~M Tomita, Haoyin Xu, Ali Geisa, Qingyang Wang, Gido~M van~de Ven,
  et~al.
\newblock Representation ensembling for synergistic lifelong learning with
  quasilinear complexity.
\newblock \emph{arXiv preprint arXiv:2004.12908}, 2020.

\bibitem[Vogelstein et~al.(2007)Vogelstein, Mallik, Vogelstein, and
  Cauwenberghs]{Vogelstein2007-wz}
R~Jacob Vogelstein, Udayan Mallik, Joshua~T Vogelstein, and Gert Cauwenberghs.
\newblock Dynamically reconfigurable silicon array of spiking neurons with
  conductance-based synapses.
\newblock \emph{IEEE Trans. Neural Netw.}, 18\penalty0 (1):\penalty0 253--265,
  January 2007.

\bibitem[Wang(2020)]{Wang2020-dy}
Jane~X Wang.
\newblock Meta-learning in natural and artificial intelligence, November 2020.

\bibitem[Wang et~al.(2020)Wang, Yao, Kwok, and Ni]{wang2020generalizing}
Yaqing Wang, Quanming Yao, James~T Kwok, and Lionel~M Ni.
\newblock Generalizing from a few examples: A survey on few-shot learning.
\newblock \emph{ACM computing surveys (csur)}, 53\penalty0 (3):\penalty0 1--34,
  2020.

\bibitem[White(1982)]{White1982-ch}
Halbert White.
\newblock {Maximum Likelihood Estimation of Misspecified Models}.
\newblock \emph{Econometrica}, 50\penalty0 (1):\penalty0 1--25, 1982.

\bibitem[Yao et~al.(2020)Yao, Zhou, Mahdavi, Li, Socher, and
  Xiong]{yao2020online}
Huaxiu Yao, Yingbo Zhou, Mehrdad Mahdavi, Zhenhui~Jessie Li, Richard Socher,
  and Caiming Xiong.
\newblock Online structured meta-learning.
\newblock \emph{Advances in Neural Information Processing Systems},
  33:\penalty0 6779--6790, 2020.

\bibitem[Yao et~al.(2022)Yao, Choi, Cao, Lee, Koh, and Finn]{yao2022wild}
Huaxiu Yao, Caroline Choi, Bochuan Cao, Yoonho Lee, Pang Wei~W Koh, and Chelsea
  Finn.
\newblock Wild-time: A benchmark of in-the-wild distribution shift over time.
\newblock \emph{Advances in Neural Information Processing Systems},
  35:\penalty0 10309--10324, 2022.

\bibitem[Ye et~al.(2022)Ye, Li, Bai, Yu, Hong, Zhou, Li, and Zhu]{ye2022ood}
Nanyang Ye, Kaican Li, Haoyue Bai, Runpeng Yu, Lanqing Hong, Fengwei Zhou,
  Zhenguo Li, and Jun Zhu.
\newblock Ood-bench: Quantifying and understanding two dimensions of
  out-of-distribution generalization.
\newblock In \emph{Proceedings of the IEEE/CVF Conference on Computer Vision
  and Pattern Recognition}, pp.\  7947--7958, 2022.

\bibitem[Ying \& Pontil(2008)Ying and Pontil]{ying2008online}
Yiming Ying and Massimiliano Pontil.
\newblock Online gradient descent learning algorithms.
\newblock \emph{Foundations of Computational Mathematics}, 8:\penalty0
  561--596, 2008.

\bibitem[Zenke et~al.(2017)Zenke, Poole, and Ganguli]{zenke2017continual}
Friedemann Zenke, Ben Poole, and Surya Ganguli.
\newblock Continual learning through synaptic intelligence.
\newblock In \emph{International conference on machine learning}, pp.\
  3987--3995. PMLR, 2017.

\end{thebibliography}
\bibliographystyle{collas2023_conference}


\end{document}